%% file: main.tex
\documentclass[10pt,twocolumn,letterpaper]{article}

\usepackage[datasets]{wacv}
\input{preamble}

\definecolor{cvprblue}{rgb}{0.21,0.49,0.74}
\usepackage[pagebackref,breaklinks,colorlinks,allcolors=cvprblue]{hyperref}

\def\wacvPaperID{930}
\def\confName{WACV}
\def\confYear{2027}

\title{When Does Resolution Help a Frozen Backbone?\\
Global Attention at Resolution Predicts Scalable Adaptation\\
for Camouflaged and Marine Animal Segmentation}

\author{Tyler Rust\\
University of Delaware\\
{\tt\small trust@udel.edu}
\and
Chandra Kambhamettu\\
University of South Florida\\
{\tt\small ckambhamettu@usf.edu}
}

\begin{document}
\maketitle

\input{sec/0_abstract}
\input{sec/1_intro}
\input{sec/2_relatedwork}
\input{sec/3_methodology}
\input{sec/4_experiment}
\input{sec/5_results}
\input{sec/6_conclusion}

{
    \small
    \bibliographystyle{ieeenat_fullname}
    \bibliography{main}
}

\end{document}

%% file: preamble.tex
\usepackage{multirow}
\usepackage{colortbl}     
\usepackage{pifont}       
\usepackage{xspace}

\definecolor{samgray}{gray}{0.92}

\newcommand{\method}{\mbox{\textsc{SALT}}\xspace}
\newcommand{\methodb}[1]{\mbox{\textsc{SALT}-#1}}   
\newcommand{\best}[1]{\textbf{#1}}

\newcommand{\rka}[1]{\textbf{\textcolor{red}{#1}}}
\newcommand{\rkb}[1]{\textbf{\textcolor{blue}{#1}}}
\newcommand{\rkc}[1]{\textbf{\textcolor{ForestGreen}{#1}}}

\newcounter{tableparts}
\renewcommand{\thetableparts}{\alph{tableparts}} 

\counterwithin*{tableparts}{table}

\renewcommand{\paragraph}[1]{\vspace{.45em}\noindent\textbf{#1.}\hspace{.5em}}

\usepackage{xspace}
\usepackage{newtxtext}

%% file: sec/0_abstract.tex
\begin{abstract}

Adapting frozen vision foundation models to fine-grained segmentation now largely depends on backbone selection. Whether the backbone applies global attention to a high-resolution token set predicts whether a low-rank adapter turns resolution into accuracy. Isotropic ViTs attend globally over the full grid and keep improving with resolution; hierarchical backbones confine early attention to local windows and pool the grid before their global stages, plateauing at lower resolutions. A controlled six-backbone study establishes the pattern, and editing the backbone points to the cause: pooling keeps the benefit, removing global attention does not. The effect is specific to low-rank adaptation. Under one fixed pipeline, \method (\textbf{S}ide-stem, \textbf{A}ttention-gated U-Net, \textbf{L}ow-rank \textbf{T}uning), one RGB-only pass on a strong isotropic backbone wins the best $S$-measure on the four data-matched camouflaged sets, and leads every marine and salient set. It reaches a new state of the art on both marine-animal benchmarks (MAS3K mIoU $0.878$).

\end{abstract}

%% file: sec/1_intro.tex
\section{Introduction}
\label{sec:intro}

\begin{figure}[t]
  \centering
  \includegraphics[width=\linewidth]{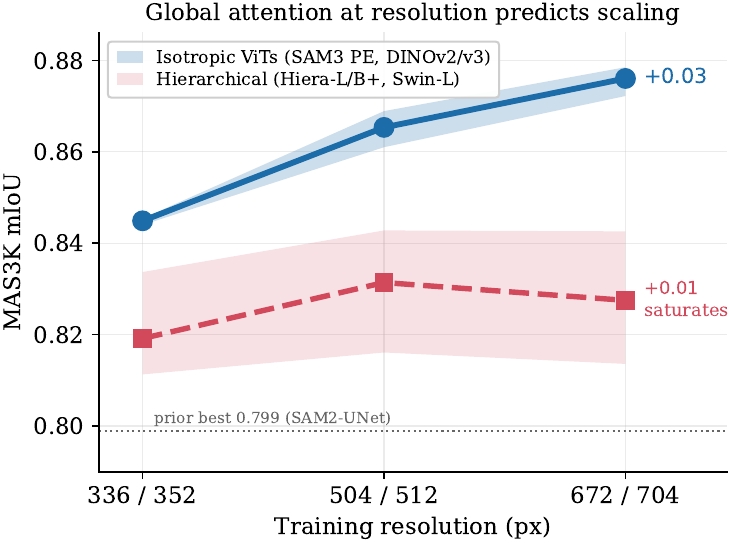}
   \caption{\textbf{Global attention at resolution, not backbone identity, predicts scaling.} On MAS3K, the three isotropic backbones (\textcolor{cvprblue}{blue}: SAM3 PE, DINOv3, DINOv2) keep turning resolution into accuracy and stay above the strongest prior (dotted, SAM2-UNet), while the three hierarchical ones (\textcolor{red}{red}: Hiera-L, Hiera-B$+$, Swin-L) saturate near $512$px, regardless of size, pretraining, or design.}
  \label{fig:teaser}
\end{figure}

Vision foundation models (VFMs) such as CLIP~\cite{radford2021learningtransferablevisualmodels}, the Segment Anything family ~\cite{kirillov2023segment,ravi2024sam2segmentimages,sam3}, and self-supervised encoders like DINOv2 and DINOv3~\cite{oquab2023dinov2,simeoni2025dinov3} have turned dense prediction into an adaptation problem where a large promptable or pretrained backbone is frozen and a small number of task parameters are learned on top~\cite{hu2021loralowrankadaptationlarge,xiong2024sam2unetsegment2makes,chen2024sam2adapterevaluatingadapting}. This regime suits fine-grained binary segmentation with scarce labels, namely camouflaged object detection (COD)~\cite{Fan_2020_CVPR_COD10K}, marine animal segmentation (MAS)~\cite{xiong2024sam2unetsegment2makes}, and salient object detection (SOD)~\cite{Zheng_2024_BiRefNet}. The shared difficulty is the decision boundary. Where a target meets a background of matched color and texture, RGB confidence collapses, and recovering the boundary demands that the backbone resolve fine structure from high-resolution input.

With many frozen backbones now available, the practical question is which one to adapt. Existing literature offers little guidance, as many adapters fix one backbone at a single resolution, confounding the pretraining contribution of the backbone with proposed modules and input resolution. We identify one structural property of the backbone that predicts how well a low-rank adapter exploits resolution, namely whether it applies global attention to a high-resolution token set. An isotropic backbone (a plain ViT such as the SAM3 Perception Encoder, DINOv3, or DINOv2) applies global self-attention over the full token grid in every block, attending a higher-resolution input globally at every depth. A hierarchical backbone, such as the Hiera encoder of SAM2, confines its early high-resolution stages to local windowed attention and pools the grid $16$--$32\times$ before global stages, and global attention therefore never operates on the high-resolution grid.

We test whether global attention predicts adaptation with our proposed decoder architecture, \method (\textbf{S}ide-stem, \textbf{A}ttention-gated U-Net, \textbf{L}oRA \textbf{T}uned): LoRA on the frozen backbone plus a high-frequency CNN side-stem U-decoder ($2.7$--$5.6$M trainable parameters, one forward pass). Six candidate backbones span two geometries, three isotropic (the SAM3 Perception Encoder, DINOv3, DINOv2) and three hierarchical (the Hiera-L and Hiera-B$+$ encoders of SAM2 and Swin-L). The result is a sharp dichotomy (Fig.~\ref{fig:teaser}, Table~\ref{tab:elastic}). Every isotropic backbone improves monotonically with resolution, while all three hierarchical backbones saturate by $512$px. The architecture is identical; the divergence is therefore a property of the selected backbone. Three controls first rule out the usual confounds. A $71$M-parameter and a $216$M-parameter Hiera both saturate at the same $512$px while isotropic backbones from $300$M to $457$M parameters keep improving (capacity); a loss that helps Hiera does not move where it saturates (objective); and self-supervised isotropic DINO backbones scale like the image--text PE (pretraining). A controlled de-pooling and attention study on two isotropic backbones then identifies the cause directly (Table~\ref{tab:mechanism}). Pooling the token grid of an isotropic backbone leaves scaling intact, while removing global attention degrades scaling on both DINOv2 and DINOv3. A native hierarchical backbone with no global attention, Swin-L, saturates like Hiera. Resolution converts into accuracy only when global attention operates on a high-resolution token set.

The fixed pipeline is also practical. On the camouflaged-detection benchmarks, a strong isotropic backbone achieves the best $S$-measure across all four datasets in a single $672$px forward pass, a lower resolution than the $1024$px baselines. We also set a new state of the art on two MAS and five SOD benchmarks.

\noindent This work is a controlled empirical study rather than a new adapter: the pipeline is held fixed and only the frozen backbone varies, and the measured effects are therefore attributable to the backbone. The contributions are:
\begin{itemize}[leftmargin=1.2em,itemsep=1pt,topsep=2pt]
  \item A \textbf{resolution-scaling dichotomy} for lightweight fine-grained adaptation. Under one fixed pipeline across six frozen backbones, isotropic backbones turn resolution into accuracy whereas hierarchical ones saturate (Fig.~\ref{fig:teaser}, Table~\ref{tab:elastic}). The effect holds on MAS and COD, where boundaries are smaller than a token, and disappears on SOD, where targets are already resolved.
  \item \textbf{A causal account of the mechanism.} Three confound controls (capacity, objective, pretraining) plus a de-pooling and windowed-attention study on two isotropic backbones (Table~\ref{tab:mechanism}) pin resolution scaling to global attention over a high-resolution token set rather than token-grid preservation. Pooling the grid does not remove scaling, removing global attention does, on both DINOv2 and DINOv3; a native Swin-L hierarchical backbone saturates like Hiera. This account is specific to low-rank adaptation: under a feature-space bottleneck adapter even the windowed Swin-L scales, global attention explains the LoRA regime rather than every adapter. Measured FLOPs and throughput make it a concrete cost--accuracy trade-off.
  \item \textbf{Practical guidance and a reproducible protocol.} The study yields a backbone-selection rule (choose an isotropic ViT, which applies global attention at high resolution) and a fully specified single-pass pipeline. We demonstrate a strong isotropic backbone reaches the best or statistically tied result across eleven benchmarks and three distinct tasks. To confirm the gains come from adaptation rather than pretrained features, an isolation study on the SAM3 PE ($0.784$ mIoU with only a trained decoder, below SAM2-UNet's $0.799$) shows adaptation drives the result.
\end{itemize}

%% file: sec/2_relatedwork.tex
\section{Related Work}
\label{sec:relatedwork}

\begin{figure*}[t]
  \centering
  \includegraphics[width=\textwidth]{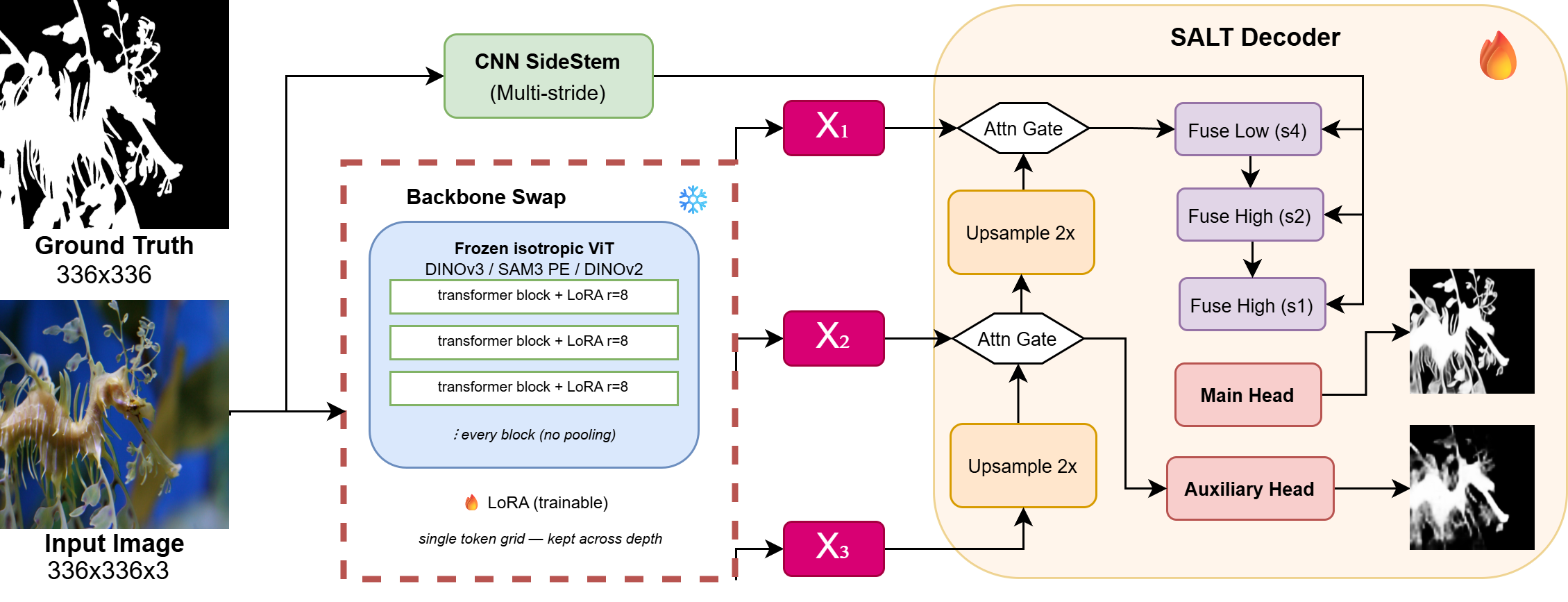}
  \caption{\textbf{The shared \method decoder.} A frozen backbone is adapted with rank-8 LoRA; multi-scale neck features are compressed by a convolutional neck, decoded by an attention-gated U-Net, and fused with a high-frequency CNN side-stem. This pipeline is applied to every backbone; only the neck differs (isotropic ViTs get a ViTDet-style feature pyramid, the hierarchical Hiera uses native stages).}
  \label{fig:arch}
\end{figure*}

\paragraph{Salient and camouflaged object detection} SOD has progressed from handcrafted saliency to deep encoder--decoders built on VGG~\cite{simonyan2015deepconvolutionalnetworkslargescale} and ResNet~\cite{he2015deepresiduallearningimage} backbones with U-Net--style skips~\cite{ronneberger2015unetconvolutionalnetworksbiomedical_unet}. Recent high-resolution methods emphasize boundary fidelity: BiRefNet~\cite{Zheng_2024_BiRefNet} uses a Swin~\cite{liu2021swin} backbone with a coarse-to-fine bilateral-reference pipeline, and FOCUS~\cite{you2025focus} unifies SOD with other foreground-segmentation tasks in a single query-based model. COD, formalized by~\cite{Fan_2020_CVPR_COD10K}, is the harder regime where targets deliberately match their surroundings; benchmarks include CAMO~\cite{ltnghia-CVIU2019_CAMO}, COD10K~\cite{Fan_2020_CVPR_COD10K}, CHAMELEON~\cite{skurowski2018animal_CHAMELEON}, and NC4K~\cite{yunqiu_cod21_nc4k}, with ZoomNeXt~\cite{Pang_2024_Zoomnext} and VSCode~\cite{luo2024vscodegeneralvisualsalient} among recent strong methods. MAS~\cite{xiong2024sam2unetsegment2makes,yan2024massam} adds underwater degradation (turbidity, caustics, low light) to the COD challenge; MAS-SAM~\cite{yan2024massam} adapts SAM with aggregated features and a pyramidal decoder for this setting. \method addresses all three (SOD, COD, and MAS) with a single RGB-only model.

\paragraph{Vision foundation models and parameter-efficient tuning}
SAM~\cite{kirillov2023segment} introduced a promptable ViT segmentation model; SAM2~\cite{ravi2024sam2segmentimages} replaced its encoder with the hierarchical Hiera transformer~\cite{ryali2023hierahierarchicalvisiontransformer} for streaming video; and SAM3~\cite{sam3} adds promptable concept segmentation on a shared Perception-Encoder (PE), an isotropic ViT-L+ that, unlike Hiera, keeps a uniform token grid across depth. The same isotropic geometry appears in the self-supervised DINOv2~\cite{oquab2023dinov2} and DINOv3~\cite{simeoni2025dinov3}, which differ from the PE in pretraining objective but not in attention pattern, a contrast we use as a control. The two geometries have been studied in isolation: the plain-ViT feature pyramid~\cite{li2022exploringvitdet} and adapters that inject multi-scale priors into an isotropic ViT for dense prediction~\cite{chen2023vitadapter}, against the staged pooling that makes Hiera efficient~\cite{ryali2023hierahierarchicalvisiontransformer}, but each fixes the input resolution and a single backbone. Reducing the deep token set, whether by spatial pooling or by merging redundant tokens~\cite{bolya2023tokenmerging}, is normally motivated by efficiency; we instead pool an isotropic backbone as a controlled probe of whether deep-grid resolution, rather than global attention, drives scaling (Sec.~\ref{sec:results-elastic}). SAM2-large exceeds $200$M parameters, and a lineage of lightweight adapters has grown around it: SAM2-Adapter~\cite{chen2024sam2adapterevaluatingadapting} inserts trainable MLPs, SAM2-UNet~\cite{xiong2024sam2unetsegment2makes} attaches a U-Net decoder, and SAM3-UNet~\cite{xiong2025sam3unet} ports the design to the SAM3 encoder. \method uses LoRA~\cite{hu2021loralowrankadaptationlarge}, adding low-rank matrices with no inference overhead. The emphasis here differs: rather than propose another adapter, we hold the pipeline fixed and vary the backbone to ask which property makes a frozen backbone adaptable. Prior adapters vary the backbone, the resolution, and the module together; sweeping six backbones under one adapter isolates the backbone's attention pattern, and an attention-editing test shows that removing global attention, not coarsening the token grid, removes the benefit (Sec.~\ref{sec:results-elastic}).

%% file: sec/3_methodology.tex
\section{Method}
\label{sec:method}

\subsection{Overview}
We study a single lightweight pipeline, \method (Fig.~\ref{fig:arch}), and apply it unchanged to every frozen backbone. It follows the encoder--decoder template of SAM2-UNet~\cite{xiong2024sam2unetsegment2makes} but freezes the backbone and adapts it with LoRA. Given an RGB image $I\in\mathbb{R}^{3\times H\times W}$, the frozen backbone with its LoRA update produces multi-scale neck features that are compressed by a light convolutional neck, decoded by an attention-gated U-Net, and fused with a high-frequency CNN side-stem to predict a binary mask $\hat S$. Only the LoRA matrices, the neck, the decoder, the side-stem, and the segmentation heads are trainable, $2.7$--$5.6$M parameters depending on the backbone (under $2\%$ of the isotropic ViTs); the variable under study is therefore the backbone rather than the adapter. The pipeline is one point in adapter space; Sec.~\ref{sec:results-elastic} tests whether the dichotomy survives a second, structurally different adapter.

\subsection{Frozen backbone with LoRA}
We instantiate \method on six frozen backbones: three isotropic ViTs, the SAM3 Perception Encoder (PE-ViT-L+, $32$ blocks, patch $14$)~\cite{sam3}, DINOv3 (ViT-L/$16$)~\cite{simeoni2025dinov3} and DINOv2 (ViT-L/$14$)~\cite{oquab2023dinov2}, and three hierarchical ones, the Hiera-L and Hiera-B$+$ encoders of SAM2~\cite{ravi2024sam2segmentimages} and Swin-L~\cite{liu2021swin}. An isotropic backbone embeds the image once into an $\tfrac{H}{p}\times\tfrac{W}{p}$ token grid that all blocks keep without spatial pooling; a hierarchical backbone pools the grid between stages. We freeze every backbone weight and inject LoRA~\cite{hu2021loralowrankadaptationlarge} into the attention projections of every block. For a frozen projection $W_0\in\mathbb{R}^{d\times k}$, LoRA learns a low-rank update
\begin{equation}
  W = W_0 + \tfrac{\alpha}{r}\,BA,\quad
  B\in\mathbb{R}^{d\times r},\;A\in\mathbb{R}^{r\times k},
\end{equation}
with a uniform rank $r{=}8$ and $\alpha{=}16$, adding no inference latency. For the isotropic ViTs a ViTDet-style simple feature pyramid~\cite{li2022exploringvitdet} turns the single token grid into three levels $X_1,X_2,X_3$ of increasing stride at $256$ channels; the native stages of Hiera and Swin already provide them. These levels are the multi-scale inputs to the decoder. This neck is the only backbone-specific component, and its asymmetry favors Hiera rather than the claim: the native Hiera stages supply genuine high-resolution early features the isotropic pyramid lacks; if the neck drove the result it would favor Hiera, which instead saturates (Sec.~\ref{sec:results-elastic}).

\input{Tables/4_Ablation}

\subsection{Compression, Side-stem, and Decoder}
Each level $X_i$ passes through a neck compression module (NCM): two $3{\times}3$ \texttt{Conv}--\texttt{BN}--\texttt{GELU} layers that reduce $256{\to}64$ channels and keep decoder cost low as the token grid grows with input resolution. The decoder is a three-stage attention-gated U-Net~\cite{ronneberger2015unetconvolutionalnetworksbiomedical_unet}: at each stage the coarser feature is bilinearly upsampled, an attention gate weights the skip feature, and the two are concatenated and passed through a \texttt{Conv}--\texttt{BN}--\texttt{GELU} double-convolution. A patch-level embedding ($14$--$16$\,px) cannot represent the finest boundaries; a lightweight CNN side-stem therefore learns features from the full-resolution input at strides $1/2/4$ and is fused into the last three decoder stages to restore high-frequency (fine boundary) detail. The branch is learned end-to-end; a hand-designed high-pass input to it does not help (Sec.~\ref{subtab:seed_band}, Table~\ref{tab:ablation}). A $1{\times}1$ head produces the mask, and an auxiliary head provides deep supervision.

\subsection{Loss}
Following~\cite{fan2020pranetparallelreverseattention_structure_loss,Wei_Wang_Huang_2020_structure_loss,xiong2024sam2unetsegment2makes} we use the structure loss, a boundary-weighted combination of binary cross-entropy and IoU,
\begin{equation}
  \mathcal{L} = \mathcal{L}^{w}_{\mathrm{BCE}} + \mathcal{L}^{w}_{\mathrm{IoU}},
\end{equation}
applied with deep supervision to the main and auxiliary outputs $S_i$ against the ground truth $G$: $\mathcal{L}_{\mathrm{total}}=\sum_i \mathcal{L}(G,S_i)$. We use no boundary or auxiliary geometric loss; none improve accuracy on these backbones (Sec.~\ref{subtab:seed_band}, Table~\ref{tab:ablation}).

\subsection{Design rationale: global attention at resolution}
\label{sec:method-elastic}
We expect isotropic backbones to be a better target for lightweight fine-grained adaptation than a hierarchical one, because the operative property is where the backbone applies global attention: an isotropic ViT attends globally over the full token grid at every depth, whereas a hierarchical backbone reaches its global stages only after pooling, so global attention never operates on the high-resolution grid. An adapter that exploits resolution should therefore scale on isotropic backbones and saturate on hierarchical ones; the cross-backbone sweep, three confound controls, and the de-pooling and windowed-attention study confirm this (Sec.~\ref{sec:results-elastic}, Tables~\ref{tab:elastic},~\ref{tab:mechanism}).

%% file: Tables/4_Ablation.tex
\begin{table}[t]
\centering
\caption{\textbf{Ablation on SAM3 PE} at $336$\,px
(MAS3K; train MAS3K\,$+$\,RMAS\,$+$\,COD10K). \emph{Top}: component ablation, then input resolution increased. \emph{Bottom}: further modules and objectives added to the baseline, each landing inside the $\pm0.002$ seed band. \method@336/@672 are two-seed means; the component and addition rows are single-seed. $\Delta$ is relative to \method@336.}
\setlength{\tabcolsep}{6pt}
\footnotesize
\begin{tabular}{l|cc}
\toprule
Variant & mIoU $\uparrow$ & $\Delta$mIoU \\
\midrule

\multicolumn{3}{l}{\refstepcounter{tableparts}\textbf{\thetableparts) Component Ablation}\label{subtab:component}} \\
\midrule
\rowcolor{samgray}
\method @336 (LoRA $+$ side-stem) & .844 & --- \\
\quad$-$ LoRA only & .835 & $-$.009 \\
\quad$-$ Side-stem only & .792 & $-$.052 \\
\quad$-$ Frozen Backbone & .784 & $-$.060 \\
\addlinespace[1ex] 

\multicolumn{3}{l}{\refstepcounter{tableparts}\textbf{\thetableparts) Input Resolution Scaling}\label{subtab:scaling}} \\
\midrule
\rowcolor{samgray}
\method @504 & .861 & $+$.017 \\
\rowcolor{samgray}
\method @672 & .872 & $+$.028 \\
\midrule

\multicolumn{3}{l}{\refstepcounter{tableparts}\textbf{\thetableparts) Additions inside the seed band:}\label{subtab:seed_band}}\\
\midrule
nested sub-patch attention & .844 & $.000$ \\
SARD structure weighting & .844 & $.000$ \\
early ViT-layer skips & .844 & $.000$ \\
hand-designed high-pass side-stem & .843 & $-$.001 \\
edge-refinement head & .842 & $-$.002 \\
\bottomrule
\end{tabular}
\label{tab:ablation}
\end{table}

%% file: sec/4_experiment.tex
\section{Experiments}
\label{sec:experiments}

\subsection{Metrics}
Following standard COD and SOD protocol we report S-measure ($S_m$)~\cite{fan2017structuremeasurenewwayevaluate}, F-measure ($F_\beta$)~\cite{Margolin_2014_CVPR_F-measure} (max $F_\beta^{x}$ for SOD, weighted $F_\beta^{w}$ for COD and MAS), mean E-measure ($E_\phi^{m}$)~\cite{fan2018enhancedalignmentmeasurebinaryforeground}, and mean absolute error ($\mathcal{M}$). For MAS we additionally report mean intersection-over-union (mIoU), the primary metric for these benchmarks.

\subsection{Datasets}
\paragraph{Marine animal segmentation (MAS)}
We train on the combined MAS3K~\cite{li2020mas3k}, RMAS~\cite{fu2023masnet}, and COD10K~\cite{Fan_2020_CVPR_COD10K} training sets and evaluate on two benchmarks: the MAS3K test set ($1{,}141$ images) and the RMAS test set ($500$ images).

\paragraph{Salient object detection (SOD)}
We train on the union of DUTS-TR~\cite{wang2017_DUTS}, HRSOD-TR~\cite{zeng2019highresolutionsalientobjectdetection_HRSOD}, and UHRSD-TR~\cite{xie2022pyramid_pgnet_uhrsd} ($17{,}092$ images) and evaluate on five standard benchmarks: DUTS-TE~\cite{wang2017_DUTS}, DUT-OMRON~\cite{yang2013saliency_DUT-OMRON}, ECSSD~\cite{Yan_2013_CVPR_ECSSD}, HKU-IS~\cite{li2015visualsaliencybasedmultiscale_HKU-IS}, and PASCAL-S~\cite{li2014secretssalientobjectsegmentation_PASCAL-S}.

\paragraph{Camouflaged object detection (COD)}
Following~\cite{Zheng_2024_BiRefNet,xiong2024sam2unetsegment2makes} we train on CAMO-TR~\cite{ltnghia-CVIU2019_CAMO}$+$COD10K-TR~\cite{Fan_2020_CVPR_COD10K} and evaluate on CAMO, COD10K, CHAMELEON~\cite{skurowski2018animal_CHAMELEON}, and NC4K~\cite{yunqiu_cod21_nc4k}.

\subsection{Implementation Details}
Each frozen backbone is initialized from its public checkpoint, the SAM3 PE-ViT-L+~\cite{sam3}, DINOv3 ViT-L/$16$~\cite{simeoni2025dinov3}, DINOv2 ViT-L/$14$~\cite{oquab2023dinov2}, SAM2.1 Hiera-L and Hiera-B$+$~\cite{ravi2024sam2segmentimages}, and Swin-L~\cite{liu2021swin}, and kept frozen; we train only LoRA ($r{=}8$, $\alpha{=}16$, attention projections), the neck, the attention-gated decoder, the side-stem, and the heads. We use AdamW~\cite{loshchilov2019decoupledweightdecayregularization_AdamW} with weight decay $1\mathrm{e}{-}4$ and cosine decay, batch size $16$, on a single RTX~5090. Augmentation is random horizontal and vertical flips and gamma jitter; inputs are square-resized. For SOD we train over $10$ epochs; for COD and MAS we train $20$ epochs at learning rate $5\mathrm{e}{-}4$. We evaluate at three input resolutions, $336/504/672$ for the ViTs and $352/512/704$ for Hiera and Swin, with no test-time augmentation and a single forward pass. The ViT sizes are multiples of $14$, so the patch-$14$ PE and DINOv2 tile them exactly into $24/36/48$ token grids; the patch-$16$ DINOv3 takes the same inputs and its stem yields $21/31/42$ grids. The two backbone families are matched by resolution tier rather than pixel-identically. Unless a cell is reported as a two-seed mean, results are single-seed point estimates; the seed-to-seed band is $\pm0.002$ mIoU/$S_m$ (Sec.~\ref{sec:ablation}), which every reported gain exceeds.

\subsection{Ablation Study}
\label{sec:ablation}
Table~\ref{tab:ablation} ablates \method on MAS3K (SAM3 PE backbone, for continuity with the frozen-backbone isolation) by removing each adaptation component. Training only the decoder on a fully frozen SAM3 (no LoRA, no side-stem) reaches just $0.784$ mIoU, $0.060$ below \method@336 and below the prior SAM2-UNet ($0.799$, Table~\ref{tab:mas}): the pretrained features are necessary but far from sufficient, and more than half of the margin over prior work comes from the adaptation, not the backbone. Among the components, low-rank adaptation dominates, removing LoRA costs $0.052$ mIoU against $0.009$ for the side-stem, which barely helps without LoRA ($0.792$ versus $0.784$); the LoRA contribution is an order of magnitude outside the $\pm0.002$ seed band, while the smaller side-stem gain is reported conservatively.


%% file: sec/5_results.tex
\section{Results}
\label{sec:results}

\subsection{Marine Animal Segmentation}
\label{sec:results-mas}
\input{Tables/Results}
Table~\ref{tab:mas} reports both MAS benchmarks. Among published methods MAS-SAM~\cite{yan2024massam} reaches $0.788$/$0.742$ mIoU (MAS3K/RMAS) and SAM2-UNet~\cite{xiong2024sam2unetsegment2makes} $0.799$/$0.738$, the strongest prior results on these two benchmarks. \methodb{DINOv3} leads every column on both: on MAS3K it reaches $0.845$ at $336$ and \best{$0.878$} at $672$ (two-seed mean), and on RMAS \best{$0.786$}, $+0.044$ over MAS-SAM; the PE and DINOv2 also reach $0.872$ and $0.877$ at $672$ (Table~\ref{tab:elastic}), confirming this is not specific to one backbone.

\subsection{Global Attention at Resolution Predicts Resolution Scaling}
\label{sec:results-elastic}
\paragraph{The dichotomy}
The question is not whether higher resolution helps but which backbones let a lightweight adapter exploit it. With \method held fixed and resolution swept across all six backbones (Table~\ref{tab:elastic}, Fig.~\ref{fig:teaser}), a sharp dichotomy appears. All three isotropic ViTs improve monotonically: DINOv3 $0.845\!\to\!0.867\!\to\!0.878$ ($+0.033$), DINOv2 $0.848\!\to\!0.869\!\to\!0.877$ ($+0.030$), and the SAM3 PE $0.844\!\to\!0.861\!\to\!0.872$ ($+0.028$). All three hierarchical backbones saturate: Hiera-L gains from $352$ to $512$ ($+0.009$) then flattens ($704$ no better than $512$), the $71$M Hiera-B$+$ behaves the same way ($0.811\!\to\!0.816\!\to\!0.814$), and the native Swin-L peaks at $512$ then declines (Table~\ref{tab:elastic}). The reported $\Delta$ is the endpoint slope (high$-$low); the sharper statement is that past the middle tier every isotropic backbone gains again (DINOv3/DINOv2/PE $+0.011/{+}0.008/{+}0.011$ mid to high) while no hierarchical one does (Hiera-L/B$+$ $+0.000/{-}0.002$, Swin-L $-0.009$ after a $+0.022$ rise to its $512$ peak): hierarchical backbones stop converting resolution at $512$, isotropic backbones keep going. The gap between the best isotropic and the best hierarchical backbone widens with resolution, from $+0.014$ to $+0.035$ mIoU. The isotropic backbones rank within a narrow margin: at $672$ (two-seed means) DINOv3 reaches $0.878$, DINOv2 $0.877$, and the PE $0.872$, leaving DINOv3 and DINOv2 tied within the seed spread. The three backbones also differ in patch size (DINOv3 is patch-$16$, DINOv2 and the PE patch-$14$), so at a matched pixel resolution their token counts differ and the ranking is token-count-confounded; the robust finding is therefore the shared scaling pattern across all three, not which isotropic backbone leads.

\paragraph{Confound controls}
The pipeline is identical; the divergence is therefore a property of the backbones, and three controls rule out the usual confounds. (i)~Capacity: the saturation persists across Hiera sizes ($71$M and $216$M both flatten at $512$, two-seed means), ruling out a capacity artifact. (ii)~Training objective: with the boundary-weighted loss that helps Hiera, Hiera-L still peaks at $512$ ($0.847$) and does not improve by $704$ ($0.846$). (iii)~Pretraining: DINOv3 and DINOv2 are self-supervised, a different objective from the image--text training of the PE, yet scale the same way. The neck does not confound it either (\S\ref{sec:method}): the ViTDet pyramid adds no information while Hiera's native stages give it genuine high-resolution early features, so the neck favors Hiera, which still saturates.

\paragraph{Isolating the cause}
We isolate the operative factor by editing the isotropic backbone directly and re-running the sweep, on DINOv2 and DINOv3 (Table~\ref{tab:mechanism}). The intuitive explanation, that hierarchical pooling coarsens the deep token grid, is wrong: pooling an isotropic backbone $2\times$ before the deep blocks does not remove the benefit. The pooled curve keeps climbing ($+0.047$ DINOv2, $+0.046$ DINOv3), rising to $+0.078$ on DINOv2 with pooling after block~2, partly a depressed low-resolution anchor not a higher ceiling ($\Delta_{\mathrm{ceil}}$, Table~\ref{tab:mechanism}); the surviving tokens still attend globally.

The factor that matters is global attention. Removing it from every block collapses scaling to $+0.001$ on DINOv2 (flat, like Hiera, against $+0.030$ unedited) and $+0.010$ on DINOv3 (against $+0.033$): near-total on DINOv2, about $70\%$ on DINOv3, whose residual is the deep grid densifying under purely local attention (at $672$, local windowing still leaves a $42\times42$ grid against $21\times21$ at $336$). The same edit validates on the true COD task (Table~\ref{tab:mechanism}, bottom): it drops the $672$ accuracy by $0.097$ mIoU and reverses scaling on the two large sets (NC4K $+0.031\!\to\!-0.011$, CAMO $+0.028\!\to\!-0.020$), a COD10K residual leaving the four-set mean halved, not flat. The native Swin-L is the confound-free necessity control: shifted-window attention, no global attention at any depth, and it saturates ($+0.013$, peaking at $512$ like Hiera), far below the isotropic $+0.033$; a reconstructed Hiera pattern inside DINOv2 saturates likewise (Table~\ref{tab:mechanism}). Resolution converts into accuracy only when global attention operates on a high-resolution token set. The no-global contrast holds on two seeds (DINOv3 and Swin endpoints, the gap at $672$ exceeding $30\times$ the band); remaining edits are single-seed, clearing it $5$--$22\times$.

The effect is task-dependent: flat on SOD, where targets are already resolved at $336$ (Sec.~\ref{sec:results-sod}), and steeper on COD than MAS ($+0.042$; Hiera saturates at $512$ on COD too; Sec.~\ref{sec:results-cod}); it pays when the task has sub-token-scale boundaries.

\textbf{Is the dichotomy adapter-specific?}
The mechanism above adapts attention, so we ask whether the dichotomy is a property of the backbone or of the adapter. Repeating the resolution sweep with a structurally different adapter, a per-token bottleneck of matched rank at every frozen block (Table~\ref{tab:adapter}), leaves the isotropic backbones scaling but flips the hierarchical one: Swin-L, which saturates under LoRA ($+0.013$), now scales ($+0.034$), and not from capacity, as it persists from bottleneck rank $16$ to $32$. A feature-space adapter lets the decoder exploit the denser high-resolution token grid even under windowed attention, so the saturation we attribute to missing global attention is specific to attention-level (LoRA) adaptation. On the SAM2 Hiera encoder the bottleneck instead collapses to the frozen-feature floor (Table~\ref{tab:adapter}), therefore Swin carries the hierarchical comparison. From these results, we scope the dichotomy and global-attention mechanism to low-rank adaptation and report the adapter dependence as a finding (Sec.~\ref{sec:conclusion}).

Does isotropic geometry cost more? Not at the useful operating point: DINOv3 at $336$ matches Hiera-L's best accuracy ($0.843$ at $512$) at lower FLOPs and higher throughput (Table~\ref{tab:elastic}), paying a premium only at $672$, for accuracy no hierarchical backbone reaches.

\input{Tables/3_Elasticity}
\input{Tables/9_Mechanism}
\input{Tables/10_Adapter}

\subsection{Salient Object Detection}
\label{sec:results-sod}
Table~\ref{tab:sod} reports \methodb{DINOv3} on five SOD benchmarks. Obtaining the best $S_m$ on all five sets, SALT-DINOv3 achieves a clear state-of-the-art on DUTS-TE and DUT-OMRON ($+0.005$ and $+0.010$ over the best published baseline) and within the seed band on PASCAL-S ($+0.002$, a tie), ahead of high-resolution methods such as BiRefNet~\cite{Zheng_2024_BiRefNet} and of SAM(2/3)-UNet~\cite{xiong2024sam2unetsegment2makes,xiong2025sam3unet}. SAM3-UNet, the nearest architectural comparison, reaches only mean $S_m$ $0.925$ against $0.933$ here. Consistent with the geometry account, SOD is where resolution scaling stops: across $336\!\to\!504\!\to\!672$ every set holds $S_m$ within the seed band (mean $0.932/0.933/0.933$), so unlike MAS and COD the extra resolution buys nothing, because salient objects are large and already resolved at $336$. Backbone choice likewise barely matters here (the PE reaches the same range as DINOv3). This is the other half of the dichotomy: the effect appears only where fine structure exists to recover.

\subsection{Camouflaged Object Detection}
\label{sec:results-cod}
Table~\ref{tab:cod} reports the terrestrial COD benchmarks. Trained identically to the MAS setting (RGB-only, single-pass), \methodb{DINOv3} at $672$ obtains the best $S_m$ on all four benchmarks (CAMO $0.927$, COD10K $0.932$, CHAMELEON $0.943$, NC4K $0.934$): a clear $+0.011$ to $+0.023$ over BiRefNet, the strongest published baseline at $1024$\,px, all from a single $672$ forward pass. Resolution scaling is steeper than on MAS: mean mIoU climbs $0.811\!\to\!0.840\!\to\!0.853$ over $336/504/672$, while the identically adapted Hiera saturates at mean mIoU $0.812$, widening the cross-backbone gap to $0.041$ mIoU.

\begin{figure}[t]
  \centering
  \includegraphics[width=\linewidth]{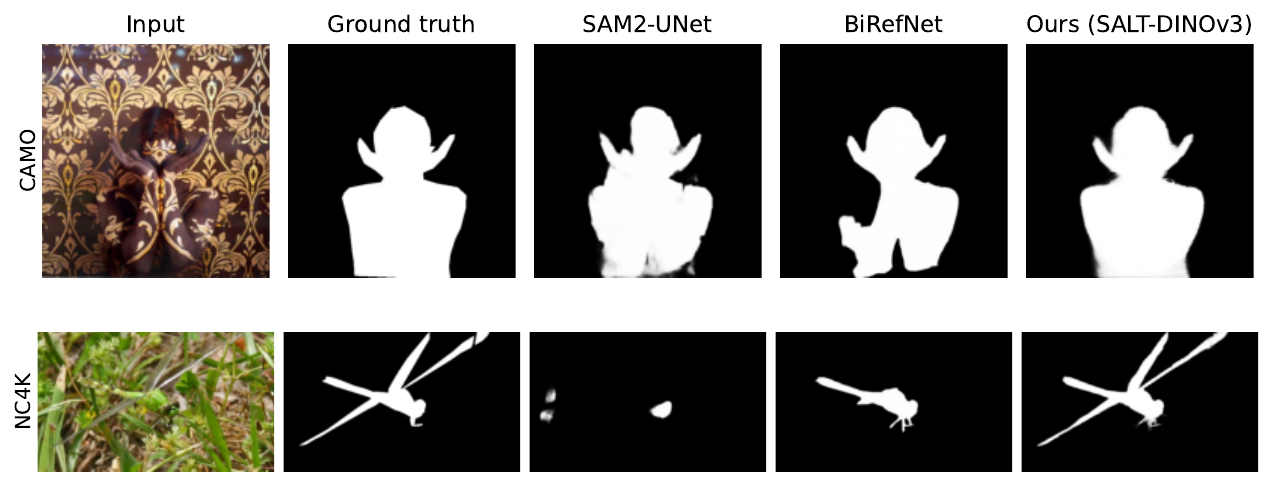}
    \caption{\textbf{Qualitative COD results} for \methodb{DINOv3} at $672$ (RGB-only, single pass) against SAM2-UNet and BiRefNet ($1024$\,px). Columns: input, ground truth, SAM2-UNet, BiRefNet, ours. Top: a texture-matched CAMO subject both baselines fragment; bottom: an NC4K dragonfly whose thin wings and tail both baselines drop. Ours recovers both (Sec.~\ref{sec:results-qual}).}
  \label{fig:qual}
\end{figure}
\subsection{Qualitative results}
\label{sec:results-qual}
Figure~\ref{fig:qual} is the qualitative counterpart to the COD gains: on a texture-matched CAMO subject and a thin-winged NC4K dragonfly, both baselines fragment the body and drop the thin structures while the RGB-only single-pass model recovers them. These are the sub-token-scale boundaries resolution scaling targets, where a saturated hierarchical backbone loses ground (Table~\ref{tab:elastic}). To confirm the cases are representative rather than cherry-picked, we scanned the full NC4K and CAMO test sets against BiRefNet's released maps (not the reported metrics, which differ): \method wins on $53\%$ of NC4K and $48\%$ of CAMO and trails on $12$--$16\%$, a mean $+0.03$ IoU. The two shown are high-margin instances of this broad advantage, not isolated wins.

%% file: Tables/Results.tex
\begin{table*}
\centering
\input{Tables/1_MAS}

\vspace{2ex}
\input{Tables/2_SOD}

\vspace{2ex}
\input{Tables/6_COD}
\end{table*}

%% file: Tables/1_MAS.tex
\caption{\textbf{Marine animal segmentation} on the MAS3K ($1{,}141$) and RMAS ($500$) test sets. Per column, the \textcolor{red}{first}, \textcolor{blue}{second}, and \textcolor{ForestGreen}{third} best are colored. Baseline numbers are quoted from the original publications.}\label{tab:mas}
\begingroup\setlength{\tabcolsep}{3pt}\scriptsize
\begin{tabular}{l|c|ccccc|ccccc}
\toprule
\multirow{2}{*}{Methods} & \multirow{2}{*}{Size}
& \multicolumn{5}{c|}{\textbf{MAS3K (1141)}}
& \multicolumn{5}{c}{\textbf{RMAS (500)}} \\
& & mIoU$\uparrow$ & $S_m\uparrow$ & $F_\beta^{w}\uparrow$ & $E_\phi^{m}\uparrow$ & $\mathcal{M}\downarrow$
& mIoU$\uparrow$ & $S_m\uparrow$ & $F_\beta^{w}\uparrow$ & $E_\phi^{m}\uparrow$ & $\mathcal{M}\downarrow$ \\
\midrule
SAM$_{23}$~\cite{kirillov2023segment} & 1024
& .566 & .763 & .656 & .807 & .059 & .445 & .697 & .534 & .790 & .053 \\
ZoomNet$_{22}$~\cite{pang2022zoomnet} & 384
& .736 & .862 & .780 & .898 & .032 & .728 & .855 & .795 & .915 & .022 \\
MASNet$_{23}$~\cite{fu2023masnet} & 352
& .742 & .864 & .788 & .906 & .032 & .731 & .862 & .801 & .920 & .024 \\
H2Former$_{23}$~\cite{he2023h2former} & 512
& .748 & .865 & .810 & .925 & .028 & .717 & .844 & .799 & .931 & .023 \\
MAS-SAM$_{24}$~\cite{yan2024massam} & 512
& .788 & .887 & .840 & .938 & .025 & .742 & .865 & .819 & .948 & .021 \\
Dual-SAM$_{24}$~\cite{zhang2024dualsam} & 512
& .789 & .884 & .838 & .933 & .023 & .735 & .860 & .812 & .944 & .022 \\
SAM2-UNet$_{26}$~\cite{xiong2024sam2unetsegment2makes} & 352
& .799 & .903 & .848 & .943 & .021 & .738 & .874 & .810 & .944 & .022 \\
\midrule
\rowcolor{samgray}
\methodb{DINOv3} & 336
& \rkc{.845} & \rkc{.932} & \rkc{.892} & \rkc{.965} & \rkc{.014} & \rkc{.758} & \rkc{.889} & \rkc{.835} & \rkc{.950} & \rkc{.020} \\
\rowcolor{samgray}
\methodb{DINOv3} & 504
& \rkb{.867} & \rkb{.939} & \rkb{.909} & \rkb{.970} & \rkb{.013} & \rkb{.776} & \rkb{.896} & \rkb{.850} & \rkb{.954} & \rkb{.019} \\
\rowcolor{samgray}
\methodb{DINOv3} & 672
& \rka{.878} & \rka{.943} & \rka{.919} & \rka{.974} & \rka{.012}
& \rka{.786} & \rka{.900} & \rka{.860} & \rka{.959} & \rka{.018} \\
\bottomrule
\end{tabular}
\endgroup

%% file: Tables/2_SOD.tex
\caption{\textbf{Salient object detection} on five benchmarks. ``TR'' is the training set ($1{=}$DUTS-TR, $2{=}$HRSOD-TR, $3{=}$UHRSD-TR). ``-'': not reported. Colors and baseline sourcing as in Table~\ref{tab:mas}.}\label{tab:sod}
\begingroup\setlength{\tabcolsep}{3pt}\scriptsize
\begin{tabular}{l|c|c|cccc|cccc|cccc|cccc|cccc}
\toprule
\multirow{2}{*}{Methods} & \multirow{2}{*}{Size} & \multirow{2}{*}{TR}
& \multicolumn{4}{c|}{\textbf{DUTS-TE (5019)}}
& \multicolumn{4}{c|}{\textbf{DUT-OMRON (5168)}}
& \multicolumn{4}{c|}{\textbf{ECSSD (1000)}}
& \multicolumn{4}{c|}{\textbf{HKU-IS (4447)}}
& \multicolumn{4}{c}{\textbf{PASCAL-S (850)}} \\
& & & $S_m$ & $F_\beta^{x}$ & $E_\phi^m$ & $\mathcal{M}$
& $S_m$ & $F_\beta^{x}$ & $E_\phi^m$ & $\mathcal{M}$
& $S_m$ & $F_\beta^{x}$ & $E_\phi^m$ & $\mathcal{M}$
& $S_m$ & $F_\beta^{x}$ & $E_\phi^m$ & $\mathcal{M}$
& $S_m$ & $F_\beta^{x}$ & $E_\phi^m$ & $\mathcal{M}$ \\
\midrule
MENet$_{23}$~\cite{Wang_2023_CVPR_MENet} & 352 & 1
& .905 & .912 & .937 & .028 & .850 & .834 & .891 & .045
& .928 & .955 & .954 & .031 & .927 & .948 & .966 & .023
& .872 & .890 & .913 & .054 \\
ICON-S$_{22}$~\cite{zhuge2022salientobjectdetectionintegrity_icon} & 352 & 1
& .917 & .886 & .954 & .025 & .869 & .804 & .900 & .043
& .941 & .936 & .966 & .023 & .935 & .925 & .968 & .022
& .885 & .854 & .924 & .048 \\
BiRefNet$_{24}$~\cite{Zheng_2024_BiRefNet} & 1024 & [1,2,3]
& .944 & .943 & .962 & .018 & .882 & .839 & .896 & .038
& - & - & - & - & - & - & - & -
& - & - & - & - \\
SAM2-UNet$_{26}$~\cite{xiong2024sam2unetsegment2makes} & 352 & 1
& .934 & - & .959 & .020 & .884 & - & .912 & .039
& .950 & - & .970 & .020 & .941 & - & \rkc{.971} & \rkc{.019}
& .894 & - & .931 & .043 \\
SAM3-UNet$_{25}$~\cite{xiong2025sam3unet} & 336 & 1
& .936 & - & - & .019 & .895 & - & - & \rkc{.034}
& .950 & - & - & .019 & .939 & - & - & .020
& .904 & - & - & \rka{.038} \\
\midrule
\rowcolor{samgray}
\methodb{DINOv3} & 336 & [1,2,3]
& \rkc{.947} & \rkc{.948} & \rkc{.969} & \rkc{.017} & \rkc{.902} & \rkc{.877} & \rkc{.925} & .034
& \rkc{.958} & \rkc{.970} & \rkc{.973} & \rkc{.017} & \rkc{.946} & \rkc{.957} & .970 & .019
& \rka{.906} & \rkb{.905} & \rkc{.939} & .040 \\
\rowcolor{samgray}
\methodb{DINOv3} & 504 & [1,2,3]
& \rka{.949} & \rka{.951} & \rka{.971} & \rka{.016} & \rkb{.904} & \rkb{.880} & \rkb{.928} & \rka{.032}
& \rkb{.960} & \rkb{.972} & \rkb{.976} & \rkb{.015} & \rkb{.947} & \rkb{.959} & \rkb{.973} & \rkb{.018}
& \rkc{.905} & \rkc{.905} & \rkb{.939} & \rkc{.040} \\
\rowcolor{samgray}
\methodb{DINOv3} & 672 & [1,2,3]
& \rkb{.948} & \rkb{.950} & \rkb{.970} & \rkb{.017} & \rka{.905} & \rka{.881} & \rka{.928} & \rkb{.032}
& \rka{.960} & \rka{.973} & \rka{.976} & \rka{.015} & \rka{.948} & \rka{.960} & \rka{.974} & \rka{.017}
& \rkb{.905} & \rka{.906} & \rka{.941} & \rkb{.039} \\
\bottomrule
\end{tabular}
\endgroup

%% file: Tables/6_COD.tex
\caption{\textbf{Camouflaged object detection} on four benchmarks. Rightmost column: mean mIoU over the four sets (``--'': baselines report no mIoU). ``-'': not reported. Colors and baseline sourcing as in Table~\ref{tab:mas}.}\label{tab:cod}
\begingroup\setlength{\tabcolsep}{3pt}\scriptsize
\begin{tabular}{l|c|cccc|cccc|cccc|cccc|c}
\toprule
\multirow{2}{*}{Methods} & \multirow{2}{*}{Size}
& \multicolumn{4}{c|}{\textbf{CAMO (250)}}
& \multicolumn{4}{c|}{\textbf{COD10K (2026)}}
& \multicolumn{4}{c|}{\textbf{CHAMELEON (76)}}
& \multicolumn{4}{c|}{\textbf{NC4K (4121)}}
& \multirow{2}{*}{\shortstack{mean\\mIoU$\uparrow$}} \\
& & $S_m\uparrow$ & $F_\beta^{w}\uparrow$ & $E_\phi^m\uparrow$ & $\mathcal{M}\downarrow$
& $S_m\uparrow$ & $F_\beta^{w}\uparrow$ & $E_\phi^m\uparrow$ & $\mathcal{M}\downarrow$
& $S_m\uparrow$ & $F_\beta^{w}\uparrow$ & $E_\phi^m\uparrow$ & $\mathcal{M}\downarrow$
& $S_m\uparrow$ & $F_\beta^{w}\uparrow$ & $E_\phi^m\uparrow$ & $\mathcal{M}\downarrow$ & \\
\midrule
SAM-Adapter$_{23}$~\cite{chen2023samadapter} & 1024
& .847 & .765 & .873 & .070 & .883 & .801 & .918 & .025
& .896 & .824 & .919 & .033 & - & - & - & - & -- \\
ZoomNeXt$_{24}$~\cite{Pang_2024_Zoomnext} & 384
& .889 & .857 & .945 & .041 & .898 & .827 & .956 & .018
& .924 & .885 & \rka{.975} & \rkc{.018} & .903 & .863 & .951 & .028 & -- \\
SAM2-UNet$_{26}$~\cite{xiong2024sam2unetsegment2makes} & 352
& .884 & .861 & .932 & .042 & .880 & .789 & .936 & .021
& .914 & .863 & .961 & .022 & .901 & .863 & .941 & .029 & -- \\
BiRefNet$_{24}$~\cite{Zheng_2024_BiRefNet} & 1024
& .904 & \rkc{.890} & \rkc{.954} & \rkc{.030} & \rkc{.913} & \rkc{.874} & \rkc{.960} & \rkb{.014}
& \rkc{.932} & \rka{.914} & - & \rka{.015} & .914 & \rkc{.894} & .953 & \rkc{.023} & -- \\
\midrule
\rowcolor{samgray}
\methodb{DINOv3} & 336
& \rkc{.918} & .881 & .953 & .030 & .909 & .838 & .953 & .017
& .924 & .871 & .957 & .021 & \rkc{.923} & .884 & \rkc{.956} & .024 & \rkc{.811} \\
\rowcolor{samgray}
\methodb{DINOv3} & 504
& \rkb{.925} & \rkb{.896} & \rkb{.957} & \rkb{.028} & \rkb{.925} & \rkb{.876} & \rkb{.967} & \rkc{.014}
& \rkb{.934} & \rkc{.896} & \rkc{.964} & .018 & \rkb{.932} & \rkb{.904} & \rkb{.963} & \rkb{.021} & \rkb{.840} \\
\rowcolor{samgray}
\methodb{DINOv3} & 672
& \rka{.927} & \rka{.903} & \rka{.959} & \rka{.027} & \rka{.932} & \rka{.891} & \rka{.971} & \rka{.012}
& \rka{.943} & \rkb{.912} & \rkb{.970} & \rkb{.016} & \rka{.934} & \rka{.909} & \rka{.964} & \rka{.020} & \rka{.853} \\
\bottomrule
\end{tabular}
\endgroup

%% file: Tables/3_Elasticity.tex
\begin{table}[t]
\centering
\caption{\textbf{Resolution sweep across six frozen backbones} under one fixed pipeline
(frozen backbone, rank-8 LoRA, side-stem decoder), MAS3K mIoU. Each cell gives
\textbf{mIoU} and, below, input\,px\,/\,GFLOPs\,/\,throughput (fvcore FLOPs;
img$\,$s$^{-1}$ at batch~8, RTX~5090). low/mid/high are the three resolution tiers;
$\Delta$ is high$-$low. Cells are two-seed means (seeds 1024/7, spread ${\leq}0.003$)
except the PE $504$ cell.}
\setlength{\tabcolsep}{3pt}
\resizebox{\columnwidth}{!}{
\begin{tabular}{l|c|c|ccc|c}
\toprule
\multirow{2}{*}{Backbone} & \multirow{2}{*}{Geometry} & \multirow{2}{*}{Params}
& \multicolumn{3}{c|}{mIoU $\uparrow$ \;/\; {\scriptsize px\,/\,GFLOPs\,/\,img$\,$s$^{-1}$}} & \multirow{2}{*}{$\Delta$} \\
& & & low & mid & high & \\
\midrule
\rowcolor{samgray}
DINOv3 ViT-L & isotropic & $300$M
& .845 & .867 & \best{.878} & \best{+.033} \\
\rowcolor{samgray}
& & & {\scriptsize 336/155/198} & {\scriptsize 504/338/97} & {\scriptsize 672/618/52} & \\
DINOv2 ViT-L & isotropic & $300$M
& .848 & .869 & .877 & +.030 \\
& & & {\scriptsize 336/197/184} & {\scriptsize 504/443/87} & {\scriptsize 672/787/44} & \\
SAM3 PE-ViT & isotropic & $457$M
& .844 & .861 & .872 & +.028 \\
& & & {\scriptsize 336/288/150} & {\scriptsize 504/769/50} & {\scriptsize 672/1153/34} & \\
\midrule
SAM2 Hiera-L & hierarchical & $216$M
& .834 & .843 & .843 & +.009 \\
& & & {\scriptsize 352/149/174} & {\scriptsize 512/258/107} & {\scriptsize 704/507/51} & \\
SAM2 Hiera-B$+$ & hierarchical & $71$M
& .811 & .816 & .814 & +.003 \\
& & & {\scriptsize 352/60/317} & {\scriptsize 512/128/162} & {\scriptsize 704/239/86} & \\
Swin-L & hierarchical & $197$M
& .813 & .835 & .826 & +.013 \\
& & & {\scriptsize 352/119/188} & {\scriptsize 512/232/98} & {\scriptsize 704/440/51} & \\
\bottomrule
\end{tabular}
}
\label{tab:elastic}
\end{table}

%% file: Tables/9_Mechanism.tex
\begin{table}[t]
\centering
\caption{\textbf{De-pooling and attention edits on two isotropic backbones.} DINOv2 and
DINOv3 are edited and re-run through the identical pipeline; MAS3K mIoU, except the bottom
DINOv3 panel which is mean mIoU over the four COD sets. The ``global attn'' column states
what each edit leaves of global attention. $\Delta$ is the scaling slope (high$-$low);
$\Delta_{\mathrm{ceil}}$ is the $672$ accuracy relative to the unedited backbone of the same
panel. Two-seed rows: DINOv2 unedited/fully-windowed/pool-12, DINOv3 unedited (MAS)/pool-12/no-global (MAS and COD), and Hiera-L (spread ${\leq}0.003$); the rest single-seed.}
\setlength{\tabcolsep}{4pt}
\resizebox{\columnwidth}{!}{
\begin{tabular}{l|c|ccc|cc}
\toprule
Configuration & {\scriptsize global attn} & low & mid & high & $\Delta$ & $\Delta_{\mathrm{ceil}}$ \\
\midrule

\multicolumn{7}{l}{\refstepcounter{tableparts}\textbf{\thetableparts)} \textit{DINOv2 (edited), MAS3K mIoU at $336/504/672$}\label{subtab:dinov2}}\\
\rowcolor{samgray}
isotropic (unedited) & full & .848 & .869 & \best{.877} & $+$.030 & --- \\
\;$+$ pool grid (after blk 12) & kept & .816 & .849 & .864 & +.047 & $-$.013 \\
\;$+$ pool grid (after blk 2) & kept & .750 & .803 & .827 & +.078 & $-$.050 \\
\;$+$ windowed-early attn & late only & .842 & .862 & .866 & +.024 & $-$.011 \\
\;$+$ Hiera pattern (window$\to$pool$\to$global) & confined & .793 & .824 & .801 & $+$.008 & $-$.076 \\
\;$+$ fully windowed (no global) & none & .799 & .806 & .800 & $+$.001 & $-$.077 \\
\midrule

\multicolumn{7}{l}{\refstepcounter{tableparts}\textbf{\thetableparts)} \textit{DINOv3 on MAS3K (edited; no-global two-seed) at $336/504/672$}\label{subtab:dinov3_mas3k}}\\
\rowcolor{samgray}
isotropic (unedited) & full & .845 & .867 & .878 & $+$.033 & --- \\
\;$+$ pool grid (after blk 12) & kept & .815 & .847 & .861 & +.046 & $-$.017 \\
\;$+$ fully windowed (no global) & none & .804 & -- & .814 & $+$.010 & $-$.065 \\
\midrule

\multicolumn{7}{l}{\refstepcounter{tableparts}\textbf{\thetableparts)} \textit{DINOv3 on COD (camocod, mean mIoU over 4 sets; no-global two-seed) at $336/504/672$}\label{subtab:dinov3_cod}}\\
\rowcolor{samgray}
isotropic (unedited) & full & .811 & .840 & .853 & $+$.042 & --- \\
\;$+$ fully windowed (no global) & none & .735 & -- & .756 & $+$.021 & $-$.097 \\
\midrule
Hiera-L (native) & confined & .834 & .843 & .843 & +.009 & --- \\
\bottomrule
\end{tabular}
}
\label{tab:mechanism}
\end{table}

%% file: Tables/10_Adapter.tex
\begin{table}[t]
\centering
\caption{\textbf{LoRA versus a bottleneck adapter.} MAS3K mIoU under two parameter-matched
adapters, rank-$8$ LoRA (adapts attention) and a per-token Houlsby bottleneck (adapts
features), single-seed. low/mid/high are the three resolution tiers; $\Delta$ is
high$-$low.}
\setlength{\tabcolsep}{5pt}
\footnotesize
\begin{tabular}{l|l|ccc|c}
\toprule
Backbone & Adapter & low & mid & high & $\Delta$ \\
\midrule
\rowcolor{samgray}
DINOv3 (iso) & LoRA & .845 & .867 & .878 & $+$.033 \\
DINOv3 (iso) & bottleneck & .845 & .865 & .876 & $+$.031 \\
\rowcolor{samgray}
DINOv2 (iso) & LoRA & .848 & .869 & .877 & $+$.030 \\
DINOv2 (iso) & bottleneck & .843 & -- & .874 & $+$.031 \\
\rowcolor{samgray}
Swin-L (hier) & LoRA & .813 & .835 & .826 & \textbf{$+$.013} \\
Swin-L (hier) & bottleneck & .813 & .837 & .848 & \textbf{$+$.034} \\
\bottomrule
\end{tabular}
\label{tab:adapter}
\end{table}

%% file: sec/6_conclusion.tex
\section{Conclusion}
\label{sec:conclusion}

We asked which property of a frozen vision backbone makes it amenable to lightweight, resolution-scalable adaptation, and answered with a controlled cross-backbone study: with \method held fixed and only the backbone varied across six candidates, whether the backbone applies global attention to a high-resolution token set predicts the outcome. Every isotropic ViT (SAM3 PE, DINOv3, DINOv2) is capable of converting input resolution into accuracy because it attends globally over the full token grid at every depth; the hierarchical backbones saturate because they confine global attention to the deepest post-pooling stages, and the native Swin-L saturates because it applies no global attention at all. Capacity, objective, and pretraining controls rule out these confounds, and a de-pooling and windowed-attention study on two isotropic backbones attributes the cause to global attention rather than the resolution of the token grid: pooling an isotropic backbone leaves scaling intact, while removing global attention degrades most of it. Measured FLOPs and throughput make the dichotomy a cost--accuracy trade-off, and an isolation study confirms the gains come from adaptation, not pretrained features. On a strong isotropic backbone \method-DINOv3 leads the best $S$-measure on every COD benchmark, is best or statistically tied across all eleven (best of two resolutions per SOD set), and sets a new state of the art on both marine-animal benchmarks, single-pass and RGB-only. To apply resolution to fine-grained boundaries under a frozen backbone, choose one that attends globally at high resolution.

\paragraph{Limitations and future work}
The attention-editing study now spans two isotropic backbones: on both DINOv2 and DINOv3, removing global attention collapses resolution scaling (to $+0.001$ and $+0.010$, against $+0.030$ and $+0.033$ unedited) while pooling the grid with global attention kept leaves it intact. The hierarchical side spans two families, Hiera at two capacities and a native Swin-L, both saturating by their mid resolution. The causal edits operate within isotropic backbones, therefore the isotropic-versus-hierarchical dichotomy itself stays correlational. The DINOv3 no-global contrast and the Swin endpoints are two-seed means, with the remaining edited rows single-seed and every delta clearing the $\pm0.002$ band by $5$ to $22\times$; and the reconstructed-hierarchical edit saturates on DINOv2 but scales on DINOv3, since the mild $2\times$ pool still leaves global attention on a growing grid, the native Swin control rather than the reconstruction carries the hierarchical claim. The band is estimated from paired seeds rather than a full confidence interval, making the within-band PASCAL-S result reported as a tie. The dichotomy is also specific to low-rank adaptation: under a per-token bottleneck adapter the isotropic backbones still scale, but the hierarchical Swin-L scales too (Sec.~\ref{sec:results-elastic}), so the global-attention mechanism explains the LoRA regime rather than every adapter. Mapping the full adapter--geometry interaction (the bottleneck leaves the SAM2 Hiera encoder at its frozen-feature floor, Hiera needs attention-level adaptation) is future work.